\documentclass[10pt, conference]{IEEEtran}
\IEEEoverridecommandlockouts
\usepackage{cite}
\usepackage{amsmath,amssymb,amsfonts}
\usepackage{algorithmic}
\usepackage{graphicx}
\usepackage{textcomp}
\usepackage{xcolor}
\usepackage[hidelinks]{hyperref}
\usepackage[font=normalsize]{caption}
\def\BibTeX{{\rm B\kern-.05em{\sc i\kern-.025em b}\kern-.08em
    T\kern-.1667em\lower.7ex\hbox{E}\kern-.125emX}}
\begin{document}

\title{Vision-Language Models on the Edge for Real-Time Robotic Perception}

\author{\IEEEauthorblockN{Sarat Ahmad, Maryam Hafeez, Syed Ali Raza Zaidi}
\IEEEauthorblockA{\textit{School of Electronic and Electrical Engineering}, \textit{University of Leeds, UK}\\
\{S.Ahmad, M.Hafeez, S.A.Zaidi\}@leeds.ac.uk}
}

\maketitle

\begin{abstract}
Vision–Language Models (VLMs) enable multimodal reasoning for robotic perception and interaction, but their deployment in real-world systems remains constrained by latency, limited onboard resources, and privacy risks of cloud offloading. Edge intelligence within 6G, particularly Open RAN and Multi-access Edge Computing (MEC), offers a pathway to address these challenges by bringing computation closer to the data source. This work investigates the deployment of VLMs on ORAN/MEC infrastructure using the Unitree G1 humanoid robot as an embodied testbed. We design a WebRTC-based pipeline that streams multimodal data to an edge node and evaluate LLaMA-3.2-11B-Vision-Instruct deployed at the edge versus in the cloud under real-time conditions. Our results show that edge deployment preserves near-cloud accuracy while reducing end-to-end latency by 5\%. We further evaluate Qwen2-VL-2B-Instruct, a compact model optimized for resource-constrained environments, which achieves sub-second responsiveness, cutting latency by more than half but at the cost of accuracy.
\end{abstract}


\begin{figure*}[!htbp]
\centering
\includegraphics[width=\textwidth, 
trim=0cm 5cm 0cm 4cm,clip
]{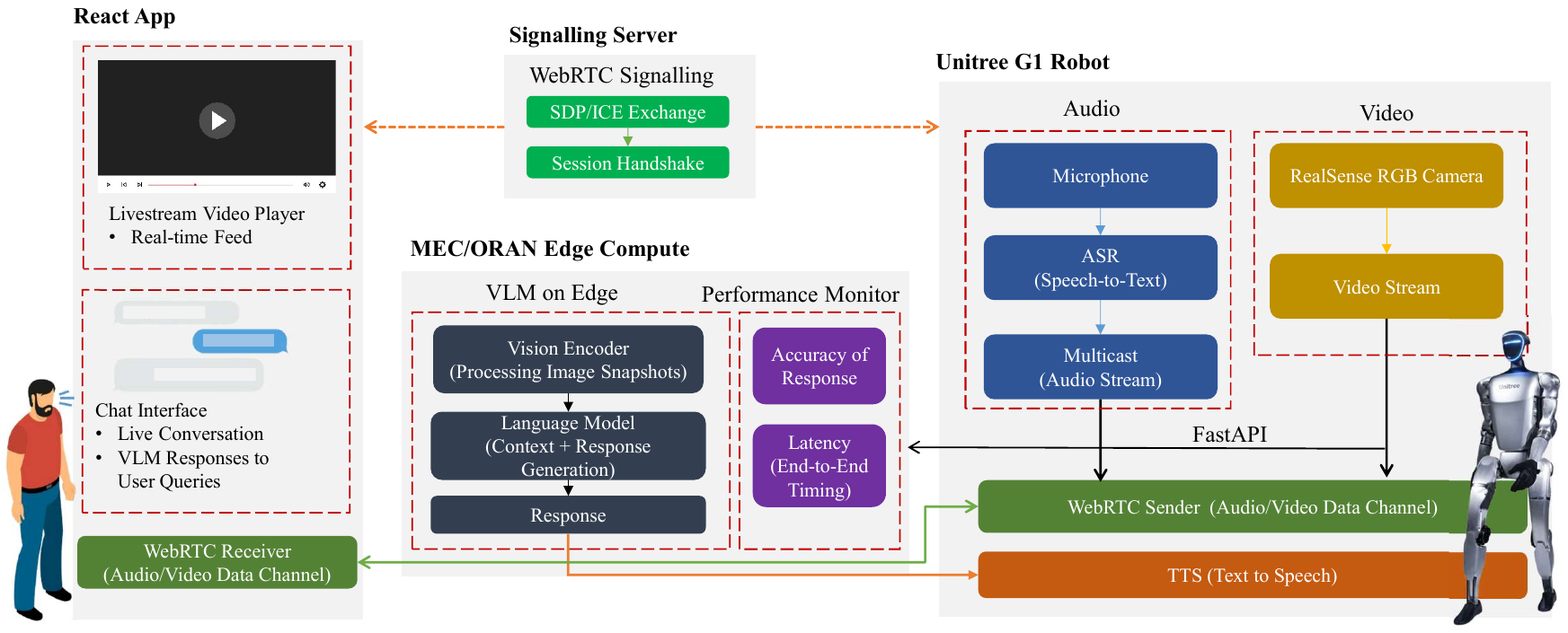}
\caption{Overview of the system architecture for edge-deployed VLM in robotic perception, illustrating the interactions among key system components.}
\label{robot_arch}
\end{figure*}

\section{Introduction}
The integration of vision and language understanding has led to the emergence of Vision–Language Models (VLMs), which couple visual perception with natural language reasoning for tasks such as image captioning, visual question answering (VQA), and embodied scene interpretation \cite{zeng2023large}. Beyond perception, VLMs show strong potential for robotic control, where their reasoning and generalization capabilities enable robots to infer human intentions, interpret emotions, and decompose complex environments into actionable steps \cite{cheng2024navila}. By processing multimodal sensor streams such as RGB, depth, and speech in real time \cite{wu2023multimodal}, VLMs enable embodied reasoning that underpins autonomous robotic capabilities, including navigation \cite{tellex2020robots}, manipulation, and human–robot interaction in both industrial and social contexts.

Despite rapid progress, realizing the full potential of VLMs in robotics depends on network infrastructures that can satisfy stringent requirements for latency, bandwidth, and privacy. Cloud-based inference, which transmits large volumes of sensitive multimodal data, is limited by scalability challenges, excessive wide-area latency, and privacy risks \cite{shuvo2022efficient}. Offloading to remote servers also creates dependency on stable connectivity, which is often unreliable in dynamic environments. Moreover, streaming raw sensory data including video, audio, and LiDAR consumes significant bandwidth and raises serious privacy concerns. These limitations highlight the need to shift inference closer to the data source. In this context, 6G-enabled edge intelligence, with compute resources embedded directly within the radio access network, offers a promising approach to support real-time perception, reasoning, and actuation in robotic systems\cite{sharshar2025vision}.

Within this context, ORAN and MEC provide a natural architectural foundation\cite{hu2015mobile}. By co-locating compute resources with radio access network components, ORAN edge nodes enable local breakout of traffic, reducing backhaul load and mitigating privacy risks. MEC further ensures the low-latency guarantees required for robotics and other interactive applications, making edge infrastructure a compelling platform for deploying VLMs in real-world human–robot interaction scenarios \cite{lin2025pushing}.

Despite this potential, few studies have empirically examined how large VLMs perform when deployed on MEC/ORAN infrastructure in real robotic systems. Existing works often rely on simulated environments\cite{dechouniotis2022edge}, virtualized edge nodes\cite{dey2016robotic}, or lightweight models, leaving open critical questions about the latency–accuracy trade-offs of deploying state-of-the-art VLMs at the edge versus the cloud. Moreover, evaluations that combine standardized benchmarks with real-world robot-collected data are rare, limiting insights into how deployment decisions affect real-time interaction.

To address this gap, we evaluate how VLMs perform under realistic robotic deployment conditions at the edge. Specifically, we examine whether large-scale models can retain their reasoning performance when migrated from cloud servers to resource-constrained edge hardware, and explore how compact models operate under strict latency constraints. This evaluation is motivated by the need to understand practical deployment trade-offs for autonomous robotics, where real-time perception and decision-making are critical.

The contributions of our work are threefold: 
\begin{enumerate} \item \textbf{System integration:} We design and implement a real-time robotic perception pipeline that deploys a VLM on an ORAN/MEC edge node, demonstrating the feasibility of large-scale multimodal inference in a realistic robotic setting.
\item \textbf{Empirical evaluation: }We present a systematic comparison of edge versus cloud deployment of LLaMA-3.2-11B-Vision-Instruct using both a standardized benchmark and a robot-collected dataset, reporting task accuracy and end-to-end latency, directly quantifying the impact of deployment location.
\item \textbf{Latency–accuracy trade-off:} We analyze the performance of a Qwen2-VL-2B-Instruct on the edge, highlighting how lightweight architectures can deliver sub-second responsiveness at reduced accuracy, thereby illustrating the design trade-offs relevant for latency-critical human–robot interaction.
\end{enumerate}

\section{Related Works}
Large Language Models (LLMs) and VLMs are increasingly integrated into robotics to enhance perception, reasoning, and human–robot interaction. SayCan\cite{ahn2022can} used PaLM to map natural-language commands to robotic actions via affordances, while PaLM-E \cite{driess2023palm} extended this to unified multimodal reasoning over vision, language, and control. More recent efforts include multimodal models for vision-and-language navigation (VLN) \cite{zitkovich2023rt}, Vision–Language–Action (VLA) models with locomotion skills for legged robots \cite{cheng2024navila}, and applications in social robotics ranging from task execution \cite{wu2023tidybot} to multi-turn, open-domain dialogue \cite{mauliana2025exploring}.

Despite these advances, deploying LLMs/VLMs in robotics remains challenging. State-of-the-art models, often with billions of parameters, are computationally intensive and exhibit high inference latency, which conflicts with the millisecond-to-second response times required for safe navigation and interaction \cite{firoozi2025foundation}. Mobile and humanoid robots are further constrained by size, weight, and power budgets, limiting access to high-end GPUs and large memory. High latency not only degrades user experience but can render systems unsafe if robots fail to react promptly.
Cloud offloading provides one solution but introduces excessive latency, dependence on network connectivity, and privacy risks \cite{lin2025pushing}. Transmitting multimodal streams (video, audio, LiDAR) to remote servers is bandwidth-intensive and unreliable in environments such as disaster zones \cite{firoozi2025foundation}. Moreover, sending raw sensory data raises significant privacy concerns \cite{lin2025pushing}. These limitations highlight the need to move inference closer to the data source.

Several strategies have been explored to mitigate these challenges. Model compression and distillation reduce memory footprint and inference cost, though often at the expense of accuracy in complex reasoning tasks \cite{lin2024awq}. Split computing and end–edge co-design distribute computation between the robot and nearby servers, lowering device-level requirements but remaining sensitive to network instability \cite{lin2025pushing}. Beyond model-level techniques, the systems community has advanced edge computing frameworks for robotics. In particular, MEC integrates compute and storage into the RAN, enabling local breakout and reducing backhaul dependence. MEC has been applied in robotics for SLAM \cite{dey2016robotic}, collaborative perception \cite{klaas2020semantic}, and real-time control \cite{tran2020autonomous,asavasirikulkij2021low}, where low-latency decision-making is essential.

While these are important advancements, real-world evaluations of edge-assisted robotic systems remain scarce. Many studies rely on virtualized edge environments without physical robots \cite{dey2016robotic}, or report limited metrics such as CPU usage while neglecting latency or task success \cite{dechouniotis2022edge}. In contrast, our work deploys a state-of-the-art VLM on a realistic MEC/ORAN edge platform using the Unitree G1 humanoid robot, and evaluates real-time performance on both standardized and robot-collected datasets in terms of latency and accuracy.

\section{System Design}
Figure \ref{robot_arch} presents an overview of the system architecture. The design integrates three key components: the Unitree G1 humanoid robot as the embodied multimodal data source, a WebRTC-based communication and control layer for real-time streaming, and the deployment of a VLM on the ORAN edge for low-latency inference and response.

\begin{figure*}[!htbp]
\centering
\includegraphics[width=\textwidth, 
trim=1.5cm 13.5cm 7.5cm 2cm,clip
]{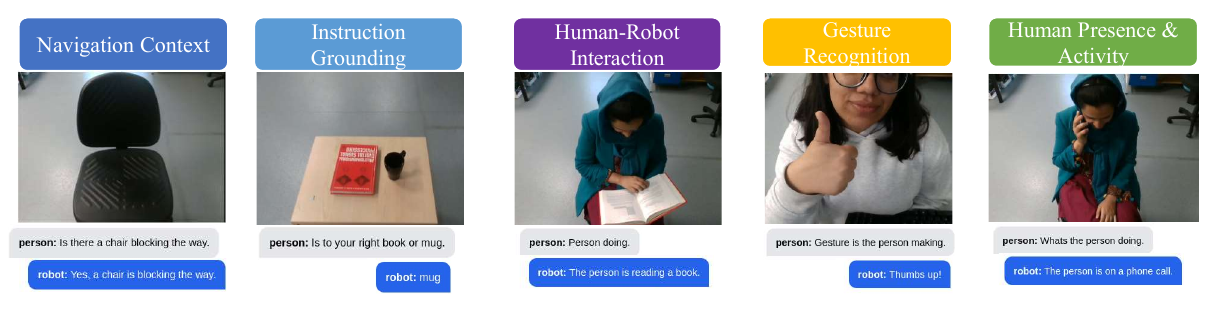}
\caption{Overview of the dataset collected with the Unitree G1 robot in a laboratory environment, comprising 200 Q\&A pairs distributed across five domains of human–robot interaction}
\label{dataset}
\end{figure*}

\subsection{Unitree G1 Robot}
We employ the Unitree G1-EDU humanoid robot \cite{g1sDKdevelopmentguide} as the embodied platform for perception and interaction. The G1 integrates multiple onboard sensors and actuators, and is recognized for its agility, precise balance control, and high-performance actuation, which enable operation in realistic laboratory environments. Within our system, the G1 also functions as a user equipment (UE) in the network, serving as the primary interface between the physical environment and the edge-deployed VLM.

For audio input, a four-microphone array captures user speech and transcribe it into text via an Automatic Speech Recognition (ASR) module. Responses generated by the VLM are converted back into speech using a Text-to-Speech (TTS) module, providing natural bidirectional audio interaction. For visual perception, the robot is equipped with an Intel RealSense D435i depth camera mounted overhead. The camera provides synchronized RGB and depth streams, allowing the system to capture both the appearance and three-dimensional structure of the environment.

In our pipeline, the G1 serves as the primary multimodal data source, continuously streaming RGB video and textual queries to the edge server and the React-based operator interface. The robot also provides a natural human–robot interaction modality: users issue queries through spoken commands or the companion application, while the robot delivers responses in spoken form. This setup enables evaluation of edge-deployed VLMs in an interactive, embodied robotic scenario under realistic operating conditions.

\subsection{React Application}
The React-based web application serves as the operator’s primary control interface, enabling real-time interaction with the robot and the edge-deployed VLM. The application renders live video streams from the Unitree G1’s RealSense camera, providing continuous visual feedback of the robot’s environment. In addition to monitoring, the interface supports an active chat module through which users can submit natural-language queries for inference on the edge server. To support transparency and system evaluation, the application also displays system-level feedback, including end-to-end latency, model response time, and accuracy of answers against annotated ground truth.

\begin{figure*}[!htbp]
\centering
\includegraphics[width=\textwidth, 
trim=0.25cm 0.2cm 0.28cm 0.2cm,clip
]{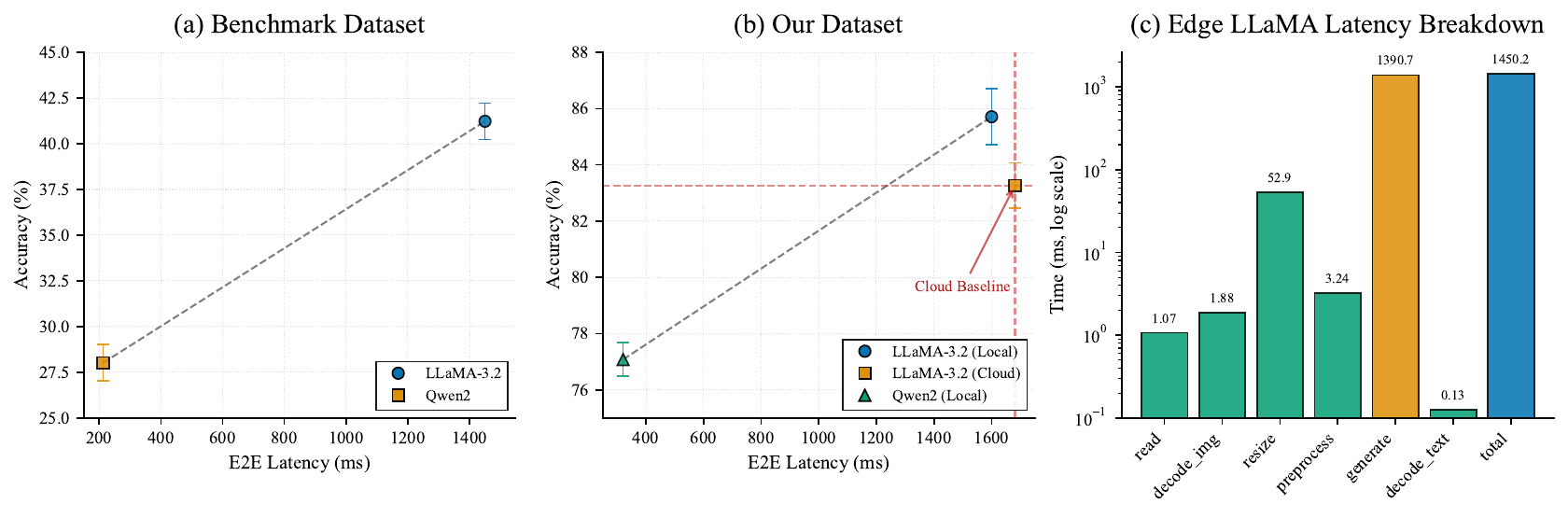}
\caption{Evaluation results: (a) Accuracy and latency of VLMs on the Robo2VLM benchmark; (b) Accuracy and latency on the robot-collected dataset; (c) End-to-end latency distributions for locally deployed LLaMA-3.2.}
\label{benchmark_plots}
\end{figure*}

\subsection{VLM Deployment on Edge}
\subsubsection{Model Selection and Configuration}
We deploy \textbf{Llama-3.2-11B-Vision-Instruct}\cite{llama-hugging-face}, an open-source VLM, to enable real-time robotic perception. The model takes an RGB frame and a textual query as input and generates a natural-language response describing or reasoning about the scene. To meet edge resource constraints, we apply 4-bit NF4 quantization with double quantization, reducing memory footprint while maintaining accuracy. Inference is optimized for latency using greedy decoding, a maximum generation length of 50 tokens, and early stopping when the model predicts an End-of-Sequence (EOS) token, ensuring concise responses without truncation.

Additionally, we also deploy \textbf{Qwen2-VL-2B-Instruct}\cite{qwen-Hugging-Face}, a compact 2B parameter VLM optimized for resource-constrained environments, on the same edge node. Including Qwen2 enables a systematic comparison between large-scale and lightweight architectures, allowing us to examine latency–accuracy trade-offs under edge constraints. This comparison is particularly relevant for highly interactive tasks such as real-time robotic perception, where responsiveness is critical.

Our edge deployment leverages an NVIDIA L4 GPU with 24 GB of VRAM, optimized for inference workloads using FP16/INT8 acceleration. This hardware profile reflects a realistic ORAN/MEC node configuration, offering a balance of efficiency and compute capacity to support real-time multimodal inference.

\subsubsection{Deployment Pipeline}
The edge node hosts a FastAPI\cite{Fastapi} inference service providing a local access point that receives frames and queries from G1 robot. Each request follows a standardized sequence: (1) image decoding and preprocessing, resize and color conversion (2) vision encoding and multimodal fusion, (3) transformer-based language generation, and (4) text decoding and return of the final response. This design prioritizes interactive, per-frame inference over batch throughput.

\subsubsection{Cloud Baseline}
For comparison, we deploy the identical Llama-3.2 model via the NVIDIA NIM cloud API \cite{NVIDIANIM}, using the same prompts and preprocessing steps as in the edge deployment. This configuration represents the traditional cloud-centric paradigm, where sensor data is transmitted over the wide-area network to remote datacenters for inference. By isolating deployment location as one of the key differences, this setup provides a baseline for quantifying the latency and accuracy benefits of ORAN edge deployment.

\subsection{Communication \& Control Layer}
To support low-latency interaction between the robot, the edge-deployed VLM, and the React application, we adopt a communication and control stack built on WebRTC\cite{webrtcGitHub}. As a browser-native, peer-to-peer framework for real-time multimedia communication, WebRTC enables the transport of video and auxiliary data streams over UDP.

\subsubsection{Signaling Server} A lightweight signaling server coordinates the initial session setup between the robot and the React application. It exchanges Session Description Protocol (SDP) offers and Interactive Connectivity Establishment (ICE) candidates, enabling peers to negotiate media capabilities and discover the optimal media path without relying on centralized relay nodes.

\subsubsection{Media Streams and Data Channels} RGB frames are transmitted over a WebRTC video track using the Secure Real-time Transport Protocol (SRTP), with adaptive bitrate control and jitter buffering to minimize end-to-end latency. Textual payloads, including ASR transcripts, operator queries, and control metadata, are exchanged via a WebRTC data channel, which provides encrypted and reliable delivery of non-media information with low latency.

\section{Evaluation}
This section introduces the evaluation metrics, describes the datasets employed, and presents discussion of the experimental results.

\subsection{Evaluation Metrics}
We evaluate system performance using two primary metrics: accuracy, which quantifies task effectiveness, and latency, which measures system responsiveness.

\subsubsection{Accuracy}  
Accuracy is defined as the proportion of model predictions that match the gold-standard answers. Following standard VQA evaluation protocols, yes/no and multiple-choice queries are scored by exact match, while short free-form responses are normalized (e.g., lowercasing, punctuation removal) prior to comparison. To isolate the effect of deployment location, we measure the accuracy of Llama-3.2 in both cloud and edge configurations. In addition, we include Qwen2 on the edge as a lightweight alternative, illustrating the latency–accuracy trade-off under constrained resources.

\subsubsection{Latency}  
End-to-end (E2E) latency is defined as the elapsed time between frame capture at the robot and receipt of the final textual response from the VLM. This measurement encompasses frame transmission, preprocessing, inference, and response decoding. By timestamping these events, we obtain precise measurements of system responsiveness. Latency distributions are reported for cloud and edge deployments of Llama-3.2, with Qwen2 on edge serving as a low-latency comparison.

\subsection{Dataset}
We evaluate our system using the Robo2VLM \cite{chen2025robo2vlm}, a large scale benchmark designed to assess VLMs in real-world robotic manipulation scenarios. Robo2VLM formulates VQA tasks from in-the-wild robot interaction datasets, pairing multimodal observations with natural-language queries. The questions span key aspects of embodied perception and control, including object recognition, spatial reasoning, manipulation intent, and outcome prediction. Each query is paired with a gold-standard answer, enabling reproducible and standardized evaluation of VLM accuracy in robotics contexts. 

In addition to the benchmark dataset, we construct a robot-collected dataset using the Unitree G1 humanoid robot in a laboratory environment. This dataset consists of 200 question–answer pairs grounded in real-time interactions. It extends beyond manipulation-focused queries to incorporate human-centered tasks such as social navigation, gesture recognition, and human presence detection\cite{shi2025hribench}. This design reflects the role of humanoid robots in navigating physical environments while engaging with human users (see Figure~\ref{dataset}). Robotic responses were collected under realistic conditions, including cluttered spaces, low-light settings, and user-driven queries issued directly in front of the robot. Importantly, the live interaction enables precise measurement of end-to-end latency by timestamping the interval between query initiation and response delivery.

\subsection{Results}
\begin{figure}[!htbp]
\centering
\includegraphics[
width=8.5cm, 
]{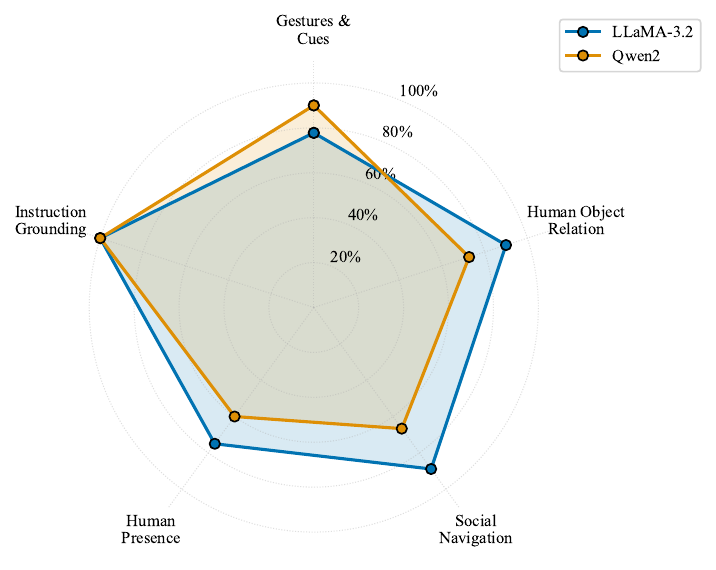}
\caption{Per-category accuracy of VLMs on the robot-collected dataset, across human–robot interaction domains.}
\label{radar_plot}
\vspace{-0.1 in}
\end{figure}

Figure \ref{benchmark_plots} summarizes the evaluation results for accuracy and end-to-end latency across benchmark and robot-collected datasets. Figure \ref{radar_plot} further breaks down accuracy by dataset category, highlighting model strengths and weaknesses across different human–robot interaction domains. In the following, we present the empirical results and discuss their implications.  

We begin with the Robo2VLM benchmark (Figure \ref{benchmark_plots}(a)), where LLaMA-3.2 achieves 41\% accuracy, while Qwen2 reaches 28.02\%, approximately 13\% lower. This performance gap reflects the difficulty of VQA tasks derived from in-the-wild robot interaction datasets, which demand fine-grained multimodal reasoning that compact models struggle to capture. 


Comparing edge to cloud deployment (Figure \ref{benchmark_plots}(b)), we find that edge LLaMA-3.2 achieves a 5\% latency reduction (1600.03 ms vs. 1685.20 ms) while also slightly improving accuracy. Although the absolute latency gain is modest, the relative improvement translates into a higher accuracy-per-millisecond ratio. For human–robot interaction, even tens of milliseconds can be significant, pushing system responsiveness closer to sub-1s thresholds for natural dialogue turn-taking, and improving safety in scenarios such as navigation, gesture recognition, and obstacle avoidance.

Qwen2 offers a different operating point, reducing latency to less than half of the cloud baseline and achieving sub-second responsiveness due to its smaller parameter size and lightweight inference path. This speed gain comes at the cost of multimodal reasoning depth and accuracy (77.08\%). Together, these results highlight a fundamental trade-off: large models such as LLaMA are \emph{performance optimal}, while compact models such as Qwen2 are \emph{efficiency optimal}. In effect, the two models trace a Pareto frontier where each is optimal under different resource and latency constraints.  

To better understand responsiveness, Figure \ref{benchmark_plots}(c) decomposes latency for locally deployed LLaMA into key components. The results show that text generation dominates total inference time, contributing over 85\% of end-to-end latency, while image decoding while image decoding and preprocessing contribute negligibly. This identifies autoregressive decoding as the primary bottleneck. System-level speedups should therefore target the generation stage, for example, through quantization, speculative decoding, or model distillation, rather than vision preprocessing.  

Task category results (Figure \ref{radar_plot}) further clarify the trade-offs. The edge-deployed LLaMA achieves balanced accuracy across all domains, demonstrating stronger multimodal grounding and reasoning. In contrast, Qwen2 performs competitively on perception-driven tasks such as human presence detection and instruction following, but lags in reasoning-intensive domains such as spatial relations and social navigation. This suggests the potential for hybrid strategies: lightweight VLMs could handle fast perceptual grounding, while larger VLMs are invoked selectively for reasoning-heavy queries.  

Finally, beyond raw latency and accuracy, edge inference brings critical system-level advantages. By avoiding the offloading of raw video and audio streams to the cloud, edge deployment reduces bandwidth consumption and mitigates privacy risks. Thus, even when latency gains are modest, edge intelligence provides robustness and trustworthiness that are essential for embodied robotic systems.  

In summary, our results demonstrate that deploying VLMs at the edge enables near-cloud accuracy while reducing latency and improving system-level efficiency. Compact models like Qwen2 deliver interactive responsiveness with reduced accuracy, while large models like LLaMA provide stronger reasoning capacity at higher computational cost. These findings underscore the value of real-world edge evaluations: they expose nuanced latency–accuracy–efficiency trade-offs and highlight adaptive deployment strategies as a path forward for VLM-driven robotics.

\section{Conclusion}
This work demonstrated the feasibility of deploying large VLMs on ORAN/MEC edge infrastructure for real-time robotic perception. Using the Unitree G1 humanoid robot as a testbed, we compared cloud-based and edge-based deployments of LLaMA-3.2-11B-Vision-Instruct, alongside a compact edge model, Qwen2-VL-2B-Instruct. Through experiments on both standardized benchmarks and a robot-collected dataset, we showed that edge deployment of LLaMA-3.2 preserves near-cloud accuracy while reducing latency, while compact models such as Qwen2 achieve sub-second responsiveness at the cost of reasoning depth. These findings establish ORAN edge infrastructure as a viable platform for multimodal AI in robotics, addressing key challenges of latency and privacy. Future work will focus on two main directions: (i) model–system co-design for latency reduction, targeting autoregressive decoding as the dominant bottleneck; and (ii) adaptive hybrid deployment strategies, where lightweight models handle fast perceptual grounding while larger models are selectively invoked for reasoning-intensive tasks.

\section*{Acknowledgment}
This research was supported by UK Research and Innovation (UKRI) through the EPSRC under two grants: the Technology Missions Fund project CHEDDAR (EP/Y037421/1), and Award UKRI851, focused on strategic decision-making and cooperation among AI agents in telecom safety and governance. This study does not involve human subjects or sensitive data, and raises no ethical or policy concerns

\bibliographystyle{IEEEtran}
\bibliography{reference}
\end{document}